\documentclass{article}
\usepackage[final]{neurips_data_2024}

\usepackage[utf8]{inputenc} %
\usepackage[T1]{fontenc}    %
\usepackage{hyperref}       %
\usepackage{url}            %
\usepackage{booktabs}       %
\usepackage{amsfonts}       %
\usepackage{nicefrac}       %
\usepackage{microtype}      %
\usepackage{xcolor}         %
\usepackage{subfig}
\usepackage{graphicx}
\usepackage{caption}
\usepackage{subcaption}

\usepackage{algorithm}
\usepackage{algpseudocode}
\usepackage{amsmath}
\usepackage{subfig}
\usepackage{fullpage}
\usepackage{enumitem}
\usepackage{xspace}

\newcommand{\sysname}{Etalon\xspace}

\newcommand{\rateMetric}{\textit{fluid token generation rate}\xspace}
\newcommand{\metric}{\textit{fluidity-index}\xspace}

\newcommand{\arxivsum}{\textit{arxiv\_summarization}\xspace}

\newcommand{\myx}{$\times$\xspace}

\newcommand{\vheading}[1]{\vspace{0.05in}\noindent\textbf{#1}}
\newcommand{\sref}[1]{\S\ref{#1}}

\newcommand{\begincompactitemize}{\begin{itemize}[noitemsep,topsep=0pt,parsep=0pt,partopsep=0pt]}

\usepackage{ifthen}
\newboolean{publicversion}
\setboolean{publicversion}{false}

\ifthenelse{\boolean{publicversion}}{
	\newcommand{\grumbler}[3]{}
        \newcommand{\jm}[1]{}
        \newcommand{\ap}[1]{}
        \newcommand{\nk}[1]{}
        \newcommand{\rr}[1]{}
        \newcommand{\sk}[1]{}
        \newcommand{\nitin}[1]{}
        \newcommand{\anmol}[1]{}
        \newcommand{\amey}[1]{}
        \newcommand{\alexey}[1]{}
}
{
\newcommand{\grumbler}[3]{\xspace\textcolor{#3}{\bf #1: #2}}
\newcommand{\jm}[1]{\grumbler{Jayashree}{#1}{magenta}}
\newcommand{\ap}[1]{\grumbler{Ashish}{#1}{violet}}
\newcommand{\nk}[1]{\grumbler{Nipun}{#1}{purple}}
\newcommand{\sk}[1]{\grumbler{sk}{#1}{purple}}
\newcommand{\alexey}[1]{\grumbler{AT}{#1}{green}}
\newcommand{\nitin}[1]{\grumbler{Nitin}{#1}{orange}}
\newcommand{\rr}[1]{\grumbler{Ram}{#1}{cyan}}
\newcommand{\anmol}[1]{\grumbler{Anmol}{#1}{cyan}}
\newcommand{\amey}[1]{\grumbler{Amey}{#1}{teal}}
}

\definecolor{commentcolor}{RGB}{0,128,0} %

\title{\sysname: Holistic Performance Evaluation Framework for LLM Inference Systems}

\author{%
Amey Agrawal$^{1*}$ Anmol Agarwal$^{1*}$ Nitin Kedia$^2$  Jayashree Mohan$^2$ Souvik Kundu$^3$ \\
\textbf{Nipun Kwatra}$^2$ \textbf{Ramachandran Ramjee}$^2$ \textbf{Alexey Tumanov}$^1$ \\\\
$^1$Georgia Institute of Technology $^2$Microsoft Research India $^3$Intel AI Lab \\
}

\begin{document}

\maketitle

\def\thefootnote{*}\footnotetext{Equal contribution. Correspondence at \texttt{ameyagrawal@gatech.edu}.}

\begin{abstract}
    Serving large language models (LLMs) in production can incur substantial costs, which has prompted recent advances in inference system optimizations. Today, these systems are evaluated against conventional latency and throughput metrics (eg. TTFT, TBT, Normalised Latency and TPOT).
    However, these metrics fail to fully capture the nuances of LLM inference, leading to an incomplete assessment of user-facing performance crucial for real-time applications such as chat and translation.
    In this paper, we first identify the pitfalls of current performance metrics in evaluating LLM inference systems. We then propose \sysname, a comprehensive performance evaluation framework  that includes \metric -- a novel metric designed to reflect the intricacies of the LLM inference process and its impact on real-time user experience.
    Finally, we evaluate various existing open-source platforms and model-as-a-service offerings using \sysname, discussing their strengths and weaknesses. \sysname is available at \href{https://github.com/project-etalon/etalon}{github.com/project-etalon/etalon}.

\end{abstract}

\section{Introduction}
\label{sec:intro}

The surge in popularity of LLMs has resulted in the proliferation of both proprietary model-as-a-service offerings~\cite{openai, azureaistudio, fireworksai, groq, anyscale} and active open-source developments aimed at optimizing LLM inference~\cite{vLLM:github, tensorrtllm:github, lightllm:github, sarathi-serve}. Given the vast array of available options, a systematic comparison of these frameworks becomes critical to ensure good user experience and cost-effective deployment.  %

Current evaluation metrics for LLM serving frameworks, such as TTFT (Time To First Token), TBT (Time Between Tokens), normalized latency, and TPOT (Time Per Output Token), fail to capture the full essence of the user experience in real-time LLM interactions. This paper demonstrates that these conventional performance metrics, while valuable, are inadequate and potentially misleading when applied to the dynamic, streaming nature of LLM inference. We argue for a more nuanced approach that considers the temporal aspects of token generation and their impact on perceived responsiveness and overall user satisfaction.

Fine-grained metrics like TTFT and TBT effectively capture latency for individual tokens and tail latency characteristics, however, they fail to represent the overall end-to-end token generation throughput. Conversely, normalized metrics such as TPOT and normalized latency attempt to measure token throughput, but fall short in identifying specific sources of user experience degradation, such as inter-token jitter or scheduling delays -- which are similar to buffering time in conventional media streaming settings. This dichotomy highlights the need for a more comprehensive evaluation framework that concisely captures the overall user experience.

To address the limitations of existing metrics, we introduce \sysname, a comprehensive framework for evaluating user-facing performance in LLM inference. At its core are two novel metrics: \metric and \rateMetric, designed to capture the nuances of real-time, streaming LLM interactions. We apply this framework to conduct an extensive performance evaluation of both open-source and proprietary LLM inference systems, revealing their strengths and weaknesses.

The design of the \metric metric is inspired by the deadline-based evaluation of periodic tasks in real-time systems~\cite{Erickson2022}. We draw an analogy between periodic tasks in the rich literature of real-time systems and the streaming token generation in LLM inference. Ideally, LLM output should maintain a smooth, consistent rate akin to media streaming platforms. However, due to various system challenges, it is difficult to maintain a constant token generation rate. \metric accounts this variability by setting token-level deadlines and measuring the fraction of deadlines met within a request. A deadline miss corresponds to a token generation stall due to delayed token generation. On the flip side, in cases of token generation exceeding the required playback rate, there's an opportunity to buffer released tokens to help mitigate future stalls and increase the fluidity. This approach enables precise and quantitative definitions of user experience constraints. For instance, it becomes possible to report the maximum load the system can sustain (capacity), subject to SLOs defined on \metric, e.g., 99\% of requests achieving a metric of $\geq$ 0.9.

\rateMetric complements \metric by determining the maximum sustainable playback rate that maintains a specified level of fluidity (e.g., \metric > 0.9). That way \rateMetric enables black-box evaluation of LLM inference systems.
Combined, these metrics provide a  holistic view of LLM inference performance that more closely aligns with real-world user experience.

We have open-sourced \sysname at \href{https://github.com/project-etalon/etalon}{github.com/project-etalon/etalon}, aiming to establish a standard for user-centric performance evaluation in the rapidly evolving landscape of LLM inference systems and frameworks.

\section{Background}
\label{sec:background}

In this section, we describe the typical LLM inference process, commonly used metrics to characterize inference performance, and an overview of open-source and proprietary inference solutions.

\subsection{LLM Inference Process}
\label{sec:background:inference-process}

There are two distinct phases in LLM inference -- a prefill phase followed by a decode phase. During prefill phase, the user's input prompt is processed and first output token is produced. Next, during the decode phase, output tokens are generated one at a time, where the token generated in one step is passed through the model to generate a new token in the next step until a special \textit{end-of-sequence} token is generated. Decode phase also requires access to KV (key and value) pairs associated with all previously processed tokens during its attention phase. Contemporary LLM inference systems store activations in KV-cache to avoid repeated re-computation during each step ~\cite{orca,fastertransformer,vllmsosp}.

\subsection{Performance Metrics for LLM Inference}
\label{sec:background:metrics}

Conventional performance metrics for LLM inference performance are the following:
\begin{itemize}[noitemsep,topsep=0pt,parsep=0pt,partopsep=0pt,leftmargin=*]
\item\vheading{TTFT} : Time To First Token (TTFT) \cite{distserve2024, llmperfdatabricks} is the latency between the request arrival and the first output token generated by the system for the request. It includes the scheduling delay (time elapsed from request arrival to start of prompt processing) and the prompt processing time.
    Minimizing TTFT is crucial for real-time interactions to maintain a responsive user experience. In contrast, longer TTFT is acceptable in offline or batch processing contexts.

\item\vheading{TBT} : Time Between Tokens (TBT) \cite{agrawal2024taming} is the latency of every subsequent token generation in the decode phase.
    This metric directly influences the perceived speed of the model by users. If we assume the average English reading speed is 250 words per minute then a TBT of roughly 6 tokens per second is required. Optimizing TBT enhances the user experience by ensuring rapid and fluid response generation.

\item\vheading{TPOT} : Time Per Output Token (TPOT) \cite{distserve2024, llmperfdatabricks} is closely related to TBT. It is the average time to generate an output token in the decode phase. It is calculated as the total decode time of a request normalized by the number of decode tokens generated.

\item\vheading{Normalized Latency} : This is defined as the total execution time of a request normalized by the number of decode tokens. It includes the scheduling delay, prompt processing time and time to generate all the decode tokens. Median Normalised Latency has been used in ~\cite{orca,vllmsosp} to compare system throughput. Lower normalised latency at a given load (queries-per-second) is desirable. 

\item\vheading{Capacity} : This is defined as the maximum request load (queries-per-second) a system can sustain while meeting certain latency targets (SLOs). It has been used in \cite{vidur, agrawal2024taming}. Higher capacity is desirable because it reduces the cost of serving.
\end{itemize}

\subsection{LLM Inference Framework Evaluation}
We now discuss what it means to evaluate the user-facing performance for LLM serving frameworks, in open-source~\cite{vLLM:github, agrawal2024taming, lightllm:github, hftgi} as well as public model offerings~\cite{openai, fireworksai, groq, azureaistudio}. 

\vheading{Open-source frameworks}. Evaluating the performance of open-source frameworks like vLLM~\cite{vLLM:github}, Sarathi-Serve~\cite{agrawal2024taming}, LightLLM~\cite{lightllm:github}, Text-Generation-Inference~\cite{hftgi}, etc. is challenging due to their numerous configurable parameters. At the same time, accurate performance assessment is crucial during deployment to determine the maximum sustainable load for a given cluster while meeting specific latency targets (SLOs).

\vheading{Proprietary model service offerings}.
Companies like OpenAI~\cite{openai}, Azure AI Studio~\cite{azureaistudio}, Fireworks AI~\cite{fireworksai}, and Groq~\cite{groq} provide model-as-a-service solutions that typically restrict end-user configurability regarding system performance. Consequently, users and developers are limited to passive performance evaluations based on their specific workload. In this constrained environment, performance comparisons across different services rely primarily on observable metrics such as latency and cost. This highlights the need for methods that can effectively guide users in selecting the most efficient and cost-effective service for their specific applications.

\section{Motivation}
\label{sec:motivation}

\subsection{Pitfalls of Existing Metrics}
\label{sec:motivation:challenges}

While the conventional latency and througput metrics described in ~\sref{sec:background:metrics} appear adequate in evaluating the performance of LLM inference systems, they fail to provide a comprehensive view of the user experience. Below, we discuss specific shortcomings identified in the current metrics.

\begin{figure}[t!]
    \centering
    \begin{minipage}{0.30\textwidth}
        \centering
        \includegraphics[width=\textwidth]{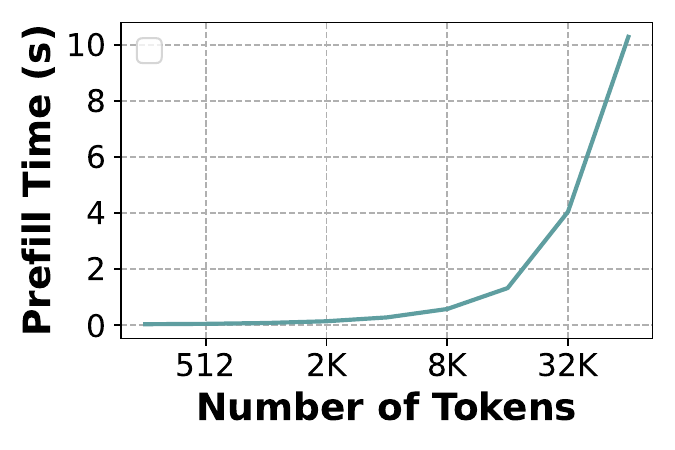}
        \caption{\small Increase in prefill latency with prompt length (Yi-34B on 2-H100) makes it infeasible to operate with fixed TTFT SLOs, especially for models with long context support.}
        \label{fig:mot:prefill}
    \end{minipage}
    \hfill
    \begin{minipage}{0.65\textwidth}
        \centering
        \subfloat[\small \centering Normalized Latency]{{\includegraphics[width=0.46\textwidth]{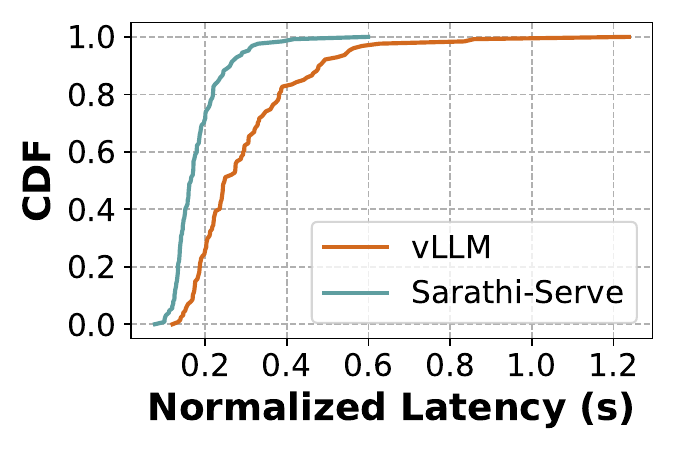} }}%
        \label{fig:mot:sd:norm_lat}
        \quad
        \subfloat[\small \centering Scheduling Delay]{{\includegraphics[width=0.46\textwidth]{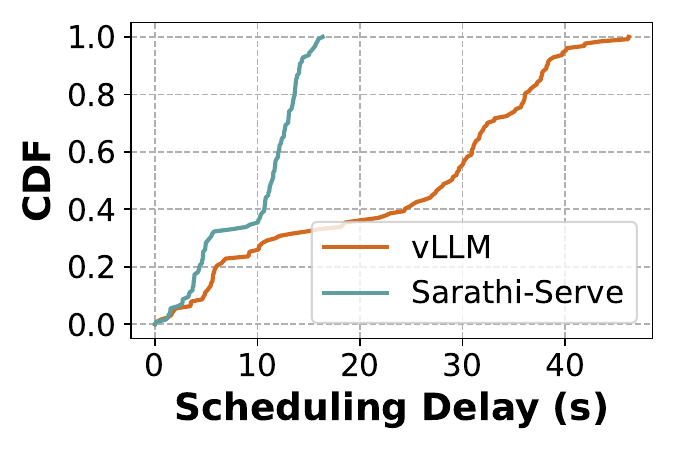} }}%
        \label{fig:mot:sd:sd}
        \caption{\small Normalized latency metric could be misleading as it obfuscates scheduling delay. On \arxivsum trace, 1.5 QPS, Yi-34B on 2-H100, while the scheduling delay is above 25s for  60\% requests in vLLM, the normalized latency only differs by few hundred ms.}
    \label{fig:mot:sd}
    \end{minipage}
\end{figure}

\vheading{Time To First Token (TTFT) is oblivious of prompt length}. TTFT, which measures prefill efficiency, includes both the scheduling delay (which depends on the system load, routing policy, batching policy \cite{orca, vllmsosp, agrawal2024taming}, etc.) and the actual prompt processing time which depends on the prompt length. Naively comparing two systems on their TTFT does not reveal the individual contribution of these components to the prefill time. Moreover, since TTFT is highly dependent on the prompt length (quadratic), as shown in ~\autoref{fig:mot:prefill}; defining a static Service Level Objective (SLO) on TTFT as a measure for user-facing responsiveness of the system is not practical. A naive alternative would be to normalize TTFT by the prompt length; but this normalizes the scheduling delay as well and would penalize shorter input requests disproportionately compared to longer ones.

\begin{figure}[t!]
\centering

 \subfloat[\small \centering Generation stalls in output tokens generation.]{{\includegraphics[width=0.28\textwidth]{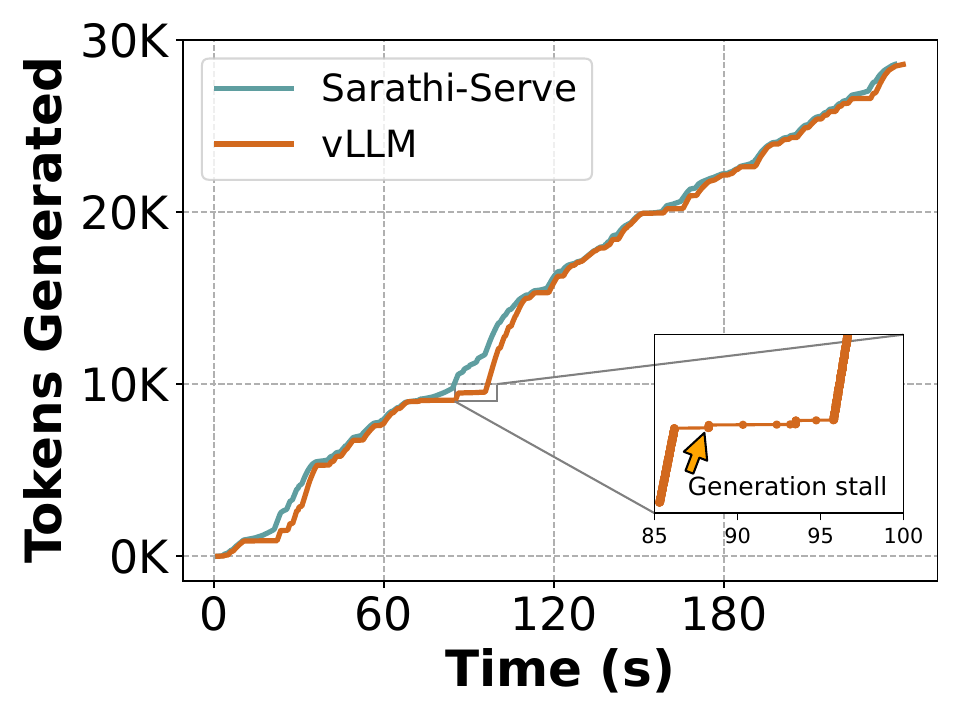} }}%
 \label{fig:mot:tbt:timeline}
\quad
\subfloat[\small \centering Token throughput derived using various latency metrics.]{{\includegraphics[width=0.32\textwidth]{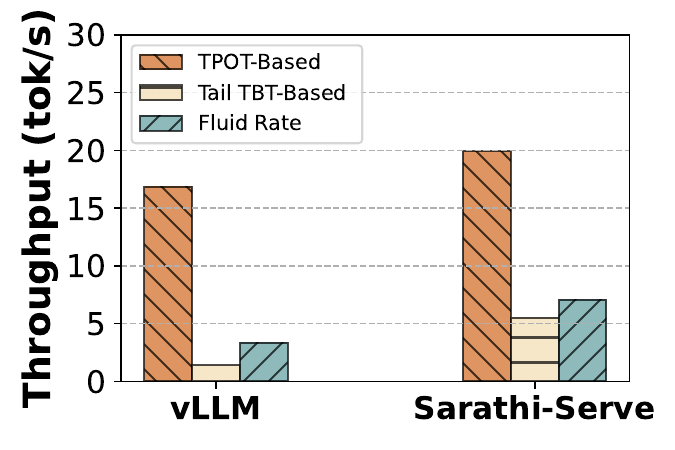} }}%
\label{fig:mot:tbt:tput}
\quad 
\subfloat[\small \centering Decode token latency distribution.]{{\includegraphics[width=0.315\textwidth]{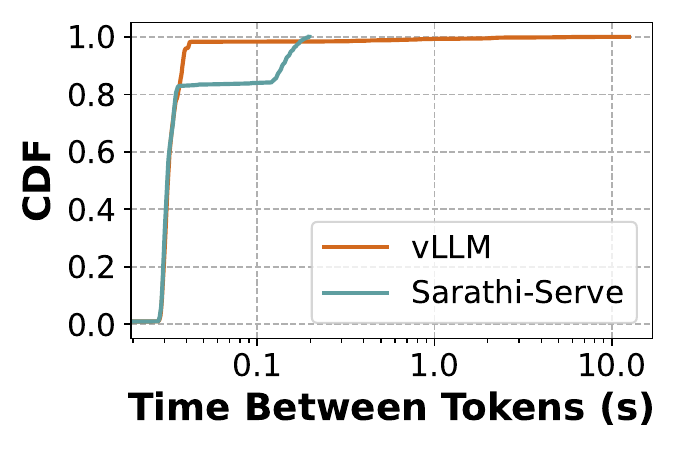} }}%
\label{fig:mot:tbt:cdf}

\caption{\small (a) Decode tokens can be intermittently stalled due to prefills from incoming requests. (b) Naively normalizing total decode latency in TPOT, hides these latency spikes and overestimates the system token throughput. (c) Simply observing tail latency does not capture the nuances in the latency distribution. P85 latency for Sarathi-Serve is higher compared to vLLM while it has much lower P99 latency. Performance evaluations with \rateMetric accounts for all these variations and provides an accurate and balanced view of system performance. Here for \rateMetric, we enforce that 99\% of the requests meet deadlines at least 90\% of the time (\metric > 0.9).}
\label{fig:mot:tbt}
\end{figure}

\vheading{Normalized latency hides scheduling delay}. Normalized latency normalizes the request end-to-end time by the total number of decode tokens (which is the visible output of the system). However, this ends up hiding specifics about metrics such as scheduling delay. For example, consider the example illustrated in \autoref{fig:mot:sd}. Here, the scheduling delay is above 25s for almost 60\% of the requests in vLLM, compared to Sarathi-Serve which has a maximum scheduling delay of 15s. However, the normalized latency for these systems differs only be a few hundred milliseconds! This is a result of the normalization by decode tokens (the median decode tokens in Arxiv-Summarization~\cite{cohan-etal-2018-discourse} is 228).

\vheading{Time Per Output Token (TPOT) and normalised latency hides jitters in token generation}. Both these metrics normalise the latency by the number of decode tokens in the request. This normalization can mask the jitters that occur as intermittent stalls during token generation. As shown in \autoref{fig:mot:tbt}a, vLLM suffers a long stall of $~10s$ (this can happen due to a long prefill request which is onboarded into the ongoing batch). While this will result in a very bad user experience, the impact of this stall on the TPOT or normalized latency metric will be numerically small due to the normalization by (typically) large number of decode tokens. Tail TBT latency, as employed by Sarathi-Serve~\cite{agrawal2024taming}, can highlight these generation stalls However, tail latency does not reveal the complete profile of magnitude and frequency of stalls at request level as shown in \autoref{fig:mot:tbt}c.

\vheading{TBT CDFs do not reveal the magnitude and temporal characteristics of stalls}. Due to the autoregressive nature of LLM inference, a delay in one token generation delays all subsequent tokens. As a result, a high tail TBT could have potentially occurred at the start of token generation, disrupting the user experience at the start itself. This is not captured by the TBT CDF. Also, as the query load increases, the frequency of stalls in systems like vLLM~\cite{vllmsosp} and Orca~\cite{orca} can go up because of prefill requests interleaved with ongoing decodes. While the tail of the TBT distribution gives some information about stalls, it does not reveal request level metrics, such as stall duration for each request, frequency and timing (say towards the start or end of its decodes) of stalls per request, etc.

\vheading{TBT fails to account for non-uniform token generation strategies}. Techniques such as speculative decoding~\cite{pmlr-v202-leviathan23a} can generate multiple decode tokens for the same request in a single iteration. Suppose 3 decode tokens $d_{i}$, $d_{i+1}$ and $d_{i+2}$ are generated in one iteration in time $T_i$. Conventionally, the TBT for $d_i$ will be $T_i$ and zero for both $d_{i+1}$ and $d_{i+2}$. Next, say the token $d_{i+3}$ is generated in time $3*T_{i}$. Naively, the TBT for $d_{i+3}$ will be attributed as $3*T_{i}$. However, the user actually saw 4 tokens generated in $4*T_{i}$ time, and the last token generation delay can be easily hidden by the user-facing client which could have shown each of the 4 tokens arriving at a uniform rate. This observation inspires our deadline based latency metric.

\vheading{Example}. Consider a scenario where an end-user, evaluates two serving systems, vLLM and Sarathi-Serve, by passively observing their response to 1000 requests. Initially, when comparing the TPOT throughput, as illustrated in \autoref{fig:mot:tbt}b, both systems appear to perform similarly. However, analysis using TBT reveals that vLLM suffers from a significantly longer tail TBT of 1 second, although it outperforms Sarathi-Serve between the 80th and 98th percentiles (see \autoref{fig:mot:tbt}c). Solely examining tail latency disproportionately penalizes vLLM, although both systems have comparable median TBTs. Thus, while TPOT downplays the discrepancies between the systems, tail-TBT overstates them. Neither metric accurately reflects the true system throughput under constraints of quality of service or user experience. Our proposed metric, \rateMetric, instead measures the actual token generate rate achievable by a system while meeting constraints on the user experience, for example, a constraint requiring a minimum percentage (e.g., 99\%) of user requests meet a token generation deadline with at least \metric of 0.9.

\subsection{Desirable Properties of Evaluation Framework}
\label{sec:motivation:desirableProps}
Having identified the pitfalls of existing metrics in evaluating LLM serving frameworks, we articulate the essential attributes of an ideal evaluation metric. First, we need an evaluation framework that is blackbox (can evaluate any API endpoint), and workload agnostic (e.g., not impacted by variance in prompt lengths in the workload).
Second, given the complexity of inferring system performance from a collection of metrics, there is a pressing need for a unified metric that not only simplifies analysis but also accurately reflects the user-facing performance of LLM serving systems, while incorporating the unique dynamics of the inference process. 
Lastly, the metric should comprehensively capture the frequency, duration, and timing of stalls within the system, addressing one of the most critical aspects affecting user experience.

\section{\sysname : Design and Implementation}
\label{sec:design}

Let us assume that we have the ideal TTFT and TBT for a given application based on some expectation on user behavior. The current TBT based SLO metrics treats each token generation independently, which may not capture the user experience well. For example, take a concrete example where the desired TBT is 100ms. Then, a system which produces 10 tokens at TBT of 10ms and the $11^{th}$ token at TBT of 150ms, will see the same TBT miss rate as a system which produces 10 tokens at TBT of 100ms and the $11^{th}$ token at TBT of 150ms. Clearly the first system is much more superior than the second, but the TBT miss rate itself does not capture that. What we propose instead is to incorporate a notion of \textit{deadline} for each token's generation. Let us analyze our example in more detail. For the first system, if the token generation started at $t=0$s, the first 10 tokens would then have been generated by $t=100$ms while the $11^{th}$ token would have been generated at $t=250$ms. If the reading speed of the user is say one token per 100ms, they would have ample time by the time they reach to the $11^{th}$ token (at $t=1000$ms). Thus, the extra delay in the $11^{th}$ token generation would not be perceived by the user. In the second system, however, the user will actually perceive a delay while reading the $11^{th}$ token. This is very similar to the case of video streaming, where TTFT corresponds to the initial delay in video playback (this includes the load times, buffering, etc.), while TBT corresponds to the delay in generation of each frame and should be below $1/fps$, where $fps$ is the video's frames per second. In the case of video playback, even if a frame is available earlier than desired, the client actively delays the frame playback to $1/fps$. Although this is not required in the case of LLM decode, the client may decide do display the tokens at TBT rate for a consistent experience even if they are available earlier. Based on this motivation, we propose a deadline based TBT acceptance rate metric, which we call \metric.

\subsection{\metric metric}
\label{sec:design:unified-deadline}

Let the desired TTFT and TBT for a given application be $D_p$ and $D_d$, respectively. Note that $D_p$ will be a function of the number of prompt/prefill tokens. We then define the \textit{deadline} for the generation of the $i^{th}$ token as $D_i = D_p + i \times D_d$. As long as all tokens are generated within their deadline, $D_i$, the user will not perceive any delay or generation stall, even if some individual tokens see a delay of more than $D_d$ between consecutive token generation. We then define a \textit{deadline-miss} as an event when the actual token generation time of the $i^{th}$ token exceeds $D_i$. Note that in the event of a \textit{deadline-miss}, depending on the length of the stall, many subsequent tokens may miss their deadlines defined as above. This can be misleading as a single stall can amount to 10s of \textit{deadline-miss}es. To account for this, if there was a \textit{deadline-miss} at the $s^{th}$ token, we reset the deadlines for all subsequent tokens to be $D_i = t_s + (i-s) \times D_d$, where $t_s$ is the actual generation time of the $s^{th}$ token, and compute the \textit{deadline-miss}es of subsequent tokens based on these refreshed deadlines.

\begin{figure}[t!]
\centering
\includegraphics[width=0.7 \textwidth]{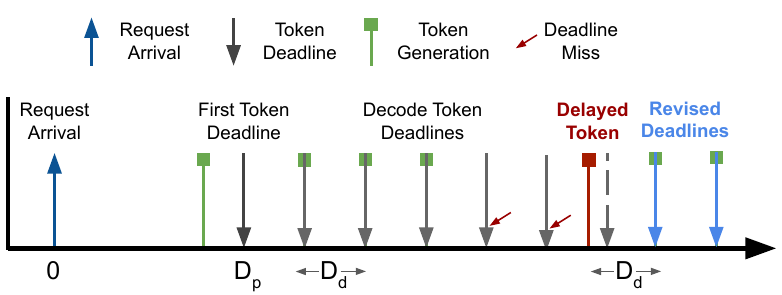}
\caption{
    \small When a request arrives in the system, \sysname sets the deadlines for all future tokens. If a token is produced before the set deadline, the slack is carried forward and serves as a buffer for future tokens. When a token arrives late, the system gets penalized for all the missed deadlines, and the subsequent deadlines are reset to account for autoregressive decoding process.
    }
\label{fig:deadline:metric}
\end{figure}

{\tiny
\begin{algorithm}
\caption{\metric computation}
\begin{algorithmic}[1]
\Require \texttt{inter\_token\_times:T},~\texttt{tbt\_deadline:D\_d},~\texttt{prefill\_tokens:P},~\texttt{scheduling\_slack:SD}
\Ensure \texttt{\metric}
\State \textcolor{commentcolor} {\texttt{\# Calculate the prefill deadline $D\_p$}}
\State \texttt{D\_p} $\gets$ \textbf{predict\_prefill\_time(}\texttt{P}\textbf{)} $+$ \texttt{SD} 
\State \texttt{total\_deadlines} $\gets 0$
\State \texttt{missed\_deadlines} $\gets 0$
\State \texttt{slack} $\gets 0$
\For{\texttt{i} $\gets 0$ \textbf{to} \texttt{length(T) - 1}}
    \State \texttt{t} $\gets$ \texttt{T[i]}
    \State \textcolor{commentcolor}{\texttt{\# Determine the appropriate deadline D}}
    \State \texttt{D} $\gets$ \texttt{D\_p} \textbf{if} \texttt{i == 0} \textbf{else} \texttt{D\_d}
    \If{\texttt{t} $\leq$ \texttt{D} $+$ \texttt{slack}}
        \State \textcolor{commentcolor}{\texttt{\# Deadline met: adjust slack and increment total deadlines}}
        \State \texttt{slack} $\gets$ \texttt{slack} $+$ \texttt{D} $-$ \texttt{t}
        \State \texttt{total\_deadlines} $\gets$ \texttt{total\_deadlines} $+$ 1
    \Else
        \State \textcolor{commentcolor}{\texttt{\# Deadline not met: calculate deadline misses}}
        \State \texttt{misses} $\gets$ (\texttt{t} $-$ \texttt{slack} $-$ \texttt{D}) $\mathbin{//}$ \texttt{D\_d} $+$ 1
        \State \texttt{missed\_deadlines} $\gets$ \texttt{missed\_deadlines} $+$ \texttt{misses}
        \State \texttt{total\_deadlines} $\gets$ \texttt{total\_deadlines} $+$ \texttt{misses}
        \State \texttt{slack} $\gets 0$
    \EndIf
\EndFor
\State \Return \texttt{(total\_deadlines - missed\_deadlines) / total\_deadlines}
\label{algo:deadline}
\end{algorithmic}
\end{algorithm}
}

\vheading{The metric}. The detailed algorithm of how \metric measures the percentage of accepted deadlines is as described in Algorithm~\ref{algo:deadline}. At a high level, we first set the deadline for all future tokens based on the prefill, decode latency target, while accounting for a small scheduling slack determined empirically. If a token is generated well before deadline, the slack is added to the future token generation, whereas in case of a missed deadline, we reset all future deadlines to start from the completion time of the token that missed the deadline as depicted in \autoref{fig:deadline:metric}.

\vheading{Picking the larget latencies}. Two important constants in the \metric metric is the target latencies for the first generated token (prefill) and subsequent (decode) tokens denoted by $D_p$ and $D_d$ above. As discussed in \sref{sec:motivation:challenges}, the prefill time increases quadratically in the size of input prompt. Therefore, setting a static value for $D_p$ is impractical as the prompt length varies significantly in production scenarios. LMSys-Chat-1M ~\cite{lmsyschat1m} has 417 median number of prefill tokens and 1418 prefill tokens at $90^{th}$ percentile. Therefore, in \sysname, we propose $D_p$ to be a function of prompt length.  We use a reasonable open-source system like vLLM~\cite{vLLM:github} as baseline to benchmark the prefill time as a function of input length by profiling the request in isolation, in the absence of any other variables like scheduling delays. We repeat this for 10 requests (tunable) and fit a curve through the observed points to obtain a $D_p$ target as a function of input length. Note that, this is a recommendation on how to set the prefill latency target; different systems may observe different performance trends due to their implementation, optimizations, and kernels used. Therefore, the process of identifying this prefill latency curve should be repeated on the system being evaluated to set practical latency targets. %

Picking the decode latency target $D_d$ is fairly intuitive and straightforward. It depends on the application being evaluated. We define three targets in \sysname -- a strict target for interactive applications like chat (25ms), medium target for medium-priority users (50ms) and a relaxed target for low-priority users (100ms). These numbers in \sysname can be picked based on production needs.

\subsection{Evaluation workflow and implementation}

\sysname, provides a standardized evaluation workflow for both proprietary and open-source LLM inference frameworks. \sysname provides two evaluation recipes as described below:

\vheading{Black-box Evaluation}. For an LLM inference API endpoint, \sysname performs black-box evaluation by hitting the server with a set of requests with diverse prompt lengths, and tracks checkpoints such as the timestamps when each output token got generated, which allows it to calculate several metrics such as TTFT, TBT, and TPOT. Moreover, given a target threshold value for \metric metric, such as -- 99\% of the requests have their \metric of at-least 0.9, \sysname infers the minimum TBT deadline that can satisfy this constraint. This allows us to obtain the maximum stable token throughput (\rateMetric) that can be served to the user in glitch-free manner.

\vheading{Capacity Evaluation}. Typically, while deploying an LLM inference service, the operator needs to determine the minimum number of GPUs required to serve the expected userbase with predefined service quality requirements. To aid this process, \sysname provides a capacity evaluation module, which evaluates a system with different request loads to identify the maximum capacity each replica can provide while meeting the \metric SLO requirements.

\vheading{Implementation}.
We create a fork of open-source LLMPerf \cite{llmperf:github} which supports calling numerous LLM APIs, specifically, OpenAI compatible APIs~\cite{openaiapi}. We extend the codebase to support interacting with open-source LLM Inference Framework vLLM~\cite{vLLM:github}, calculating existing metrics discussed in ~\sref{sec:background:metrics}, \metric metric proposed in ~\sref{sec:design:unified-deadline} and capacity search.

\section{Evaluation}
\label{sec:eval}
In this section we demonstrate the effectiveness of \sysname in holistically evaluating the performance of different LLM inference systems both open source frameworks and proprietary offerings. %

\subsection{Evaluating Public APIs}
\begin{figure}[t!]
\centering

\subfloat[\small \centering Token Generation Rate.]{{\includegraphics[width=0.45\textwidth, trim=13cm 0cm 0cm 0cm, clip]{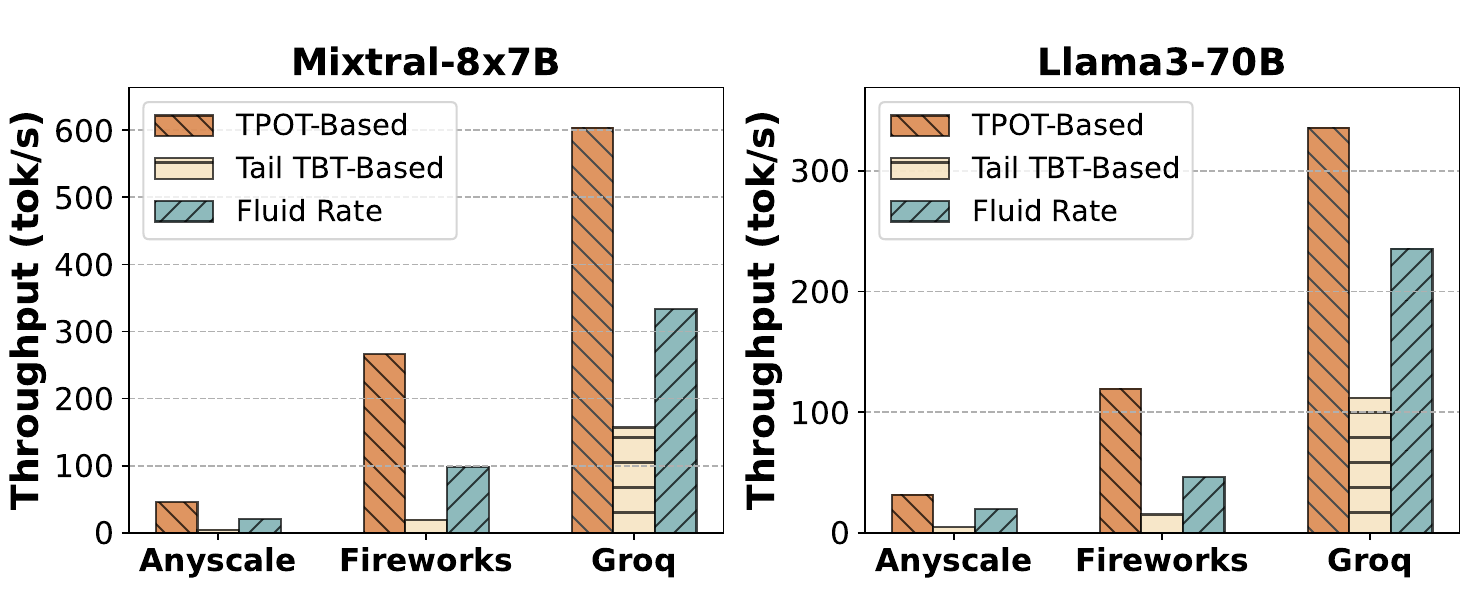} }}%
\label{fig:api:token-rate}
\quad
\subfloat[\small \centering TBT Latency Distribution.]{{\includegraphics[width=0.42\textwidth, trim=13cm 0cm 0cm 0cm, clip]{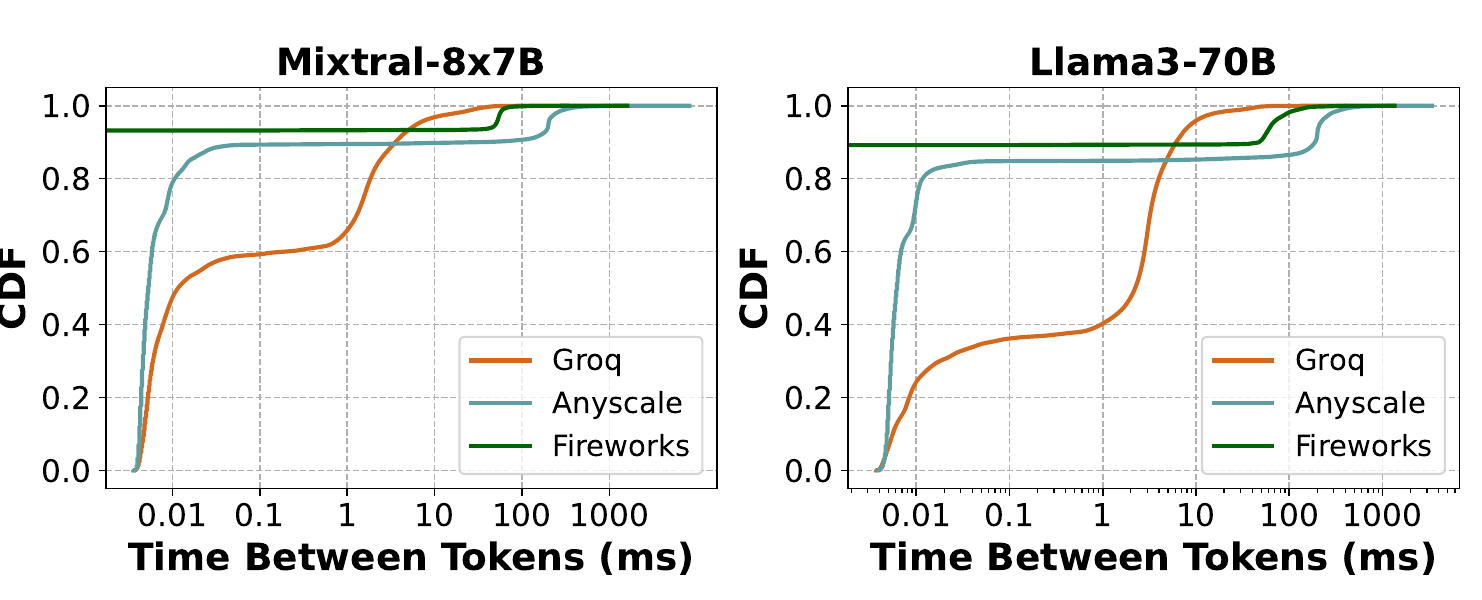} }}%
\label{fig:api:tbt-latency}
\quad
\subfloat[\small \centering TTFT Distribution.]{{\includegraphics[width=0.45\textwidth, trim=13cm 0cm 0cm 0cm, clip]{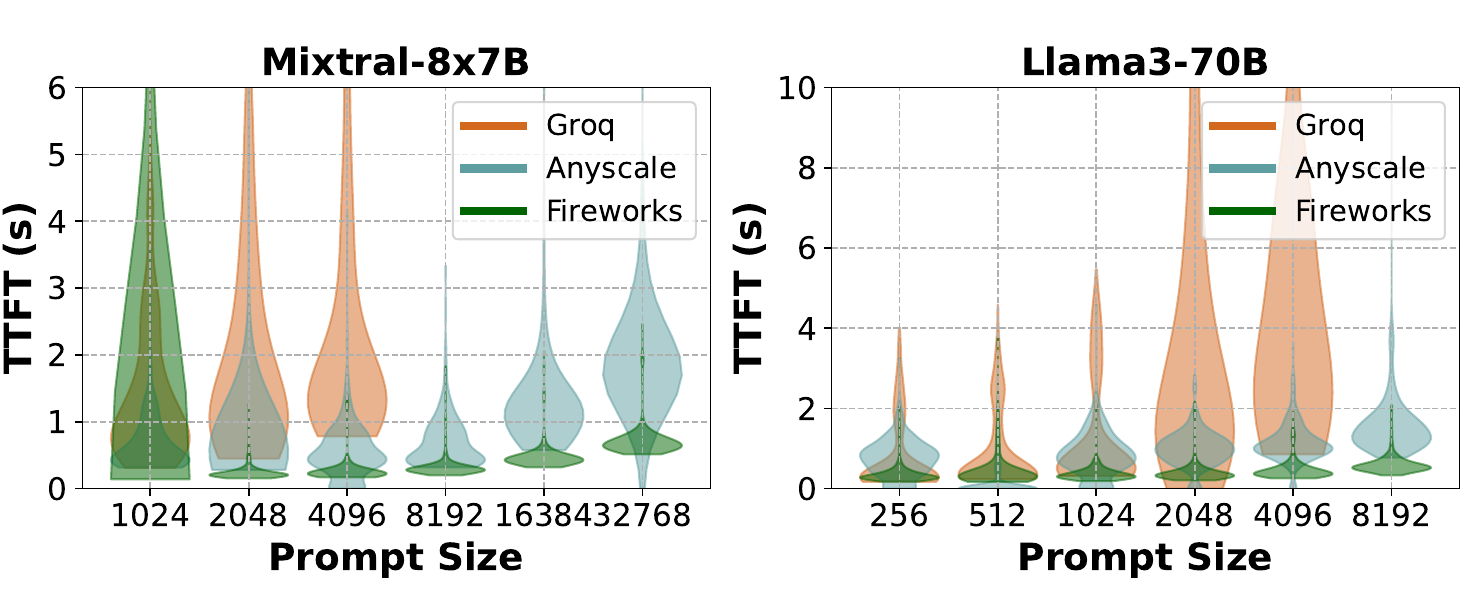} }}%
\label{fig:api:ttft}
\quad
\subfloat[\small \centering Acceptance rate as a function of TBT target.]{{\includegraphics[width=0.42\textwidth, trim=13cm 0cm 0cm 0cm, clip]{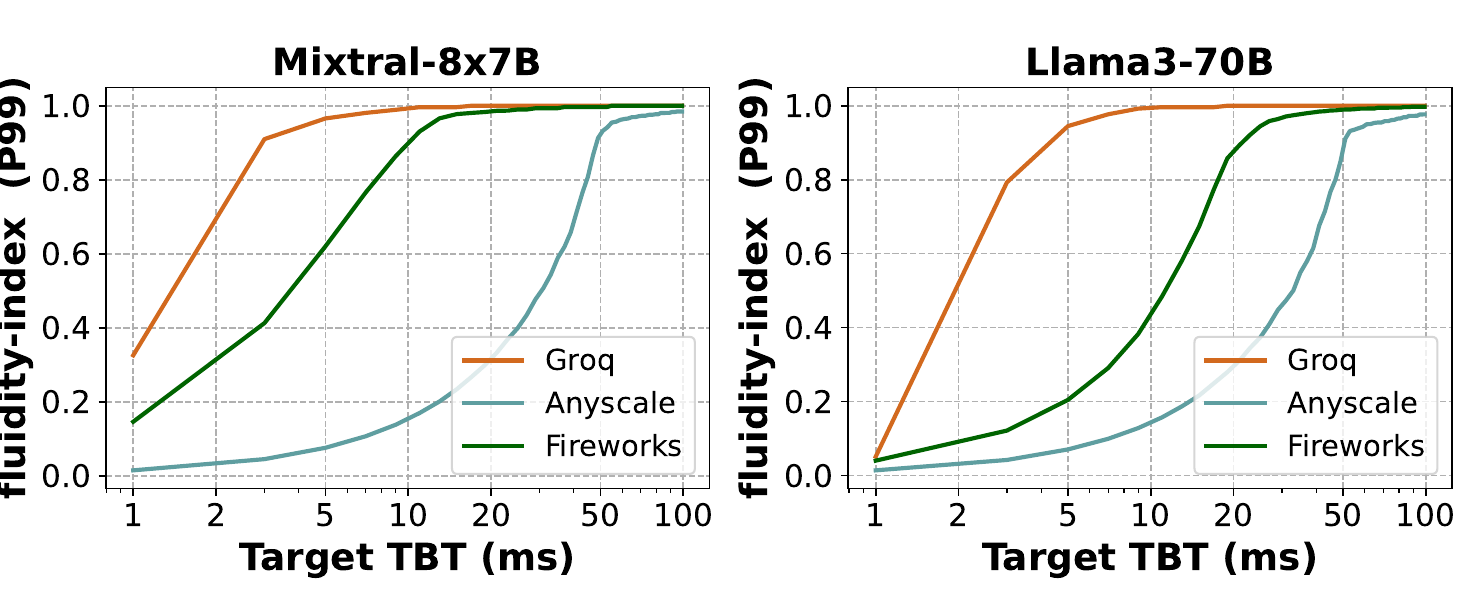} }}%
\label{fig:api:deadline}

\caption{\small Evaluation of proprietary serving offerings for Mixtral-8x7B and Llama3-70B performed over duration of 24 hrs. (a) shows the token throughput as estimated by different decode latency metrics, (b) presents the overall decode latency distribution across all requests, (c) shows the TTFT for different prompt lengths and (d) provides a full characterization of the system by showing the \metric as a function of target TBT.}
\label{fig:api}
\end{figure}

In this section, we demonstrate the effectiveness of \sysname to benchmark public API endpoints.   %
\sysname performs black-box analysis on these systems to characterize their performance under various  configurations. We evaluate three proprietary systems with API-only access: Anyscale \cite{anyscale}, Groq \cite{groq}, and Fireworks \cite{fireworksai}, across two models -- a dense model LLaMA3-70B\cite{llama3blog}, and a MoE model Mixtral-8x7B~\cite{jiang2024mixtral}. We use a custom workload with varying prefill length (between 256 and 8k and maximum tokens to generate set to 256. Since the performance of public APIs can change throughout the day depending on request traffic and other factors, we run \sysname once every hour for 24 hours to accommodate varying nature of traffic throughout the day. %

\textbf{Results.} \autoref{fig:api}a plots the throughput of the three systems using three metrics -- TPOT, tail TBT, and \metric. For the first two, we plot the inverse of the observed mean TPOT across all requests, and the inverse of the 99th percentile TBT. For \rateMetric, we find the minimum TBT SLO $D_d$ such that 99\% percent of the requests have \metric of at-least 0.9. The inverse of this is the \rateMetric.

We observe that Groq has the highest throughput of 600 tokens/s based on TPOT, a value that service providers oftentimes report. However, the tail TBT metric shows a 4\myx lesser throughput potentially due to jitters and stalls between decode iterations. The former is too relaxed and ignores generation stalls while the latter over penalizes the tail latency spikes. The throughput computed using \rateMetric lays a fair ground and shows the throughput that the system can sustain while providing a good user experience.

We make some interesting observations from the TBT distribution in ~\autoref{fig:api}b. First, in Fireworks~\cite{fireworksai}, roughly 90\% of the decodes arrive together; hints at the potential use of speculative decoding. Second, all the offerings exhibit the S-shaped curve -- indicative of prefill-decode interference due to long context length requests.
Next, the TTFT distributions in \autoref{fig:api}c shows that Fireworks consistently has the lowest TTFT as well as variance for all prompt lengths. %
The minimum TTFT for Anyscale overlaps with Fireworks indicating that their system can achieve similar prompt processing efficiency. However, the wide spread of TTFT in Anyscale indicates potentially high scheduling delays either due to high load or underprovisioning. Unlike decodes, Groq has the worst prefill efficiency. %

Finally, the deadline miss rate plots in \autoref{fig:api}d clearly highlight the differences in TBT across the three systems. Drawing a horizontal line at a desired miss rate (say 10\%), we see that for both Mixtral-8x7B and LLaMA3-70B,  Groq~\cite{groq} has the best TBT, followed by Fireworks and Anyscale. While this was difficult to interpret in \autoref{fig:api}b, \metric highlights this difference.

\subsection{Evaluating Open Source Systems}

 We now demonstrate the effectiveness of \sysname in setting SLOs for deployment operators and capacity planning. We evaluate vLLM~\cite{vLLM:github} and Sarathi-Serve \cite{agrawal2024taming} (via vLLM with \textit{chunked-prefill} feature turned on). on LLaMA3-8B \cite{llama3blog, touvron2023llama} on a H100. We use rope-scaling to support prompt token lengths longer than 8192. Requests are randomly sampled from the Arxiv-Summarization \cite{cohan-etal-2018-discourse} dataset that represents long context workloads.

\begin{figure}[t!]
\centering

\subfloat[\small \centering System Capacity.]{{\includegraphics[width=0.28\textwidth, trim=12.2cm 0cm 0cm 1cm, clip]{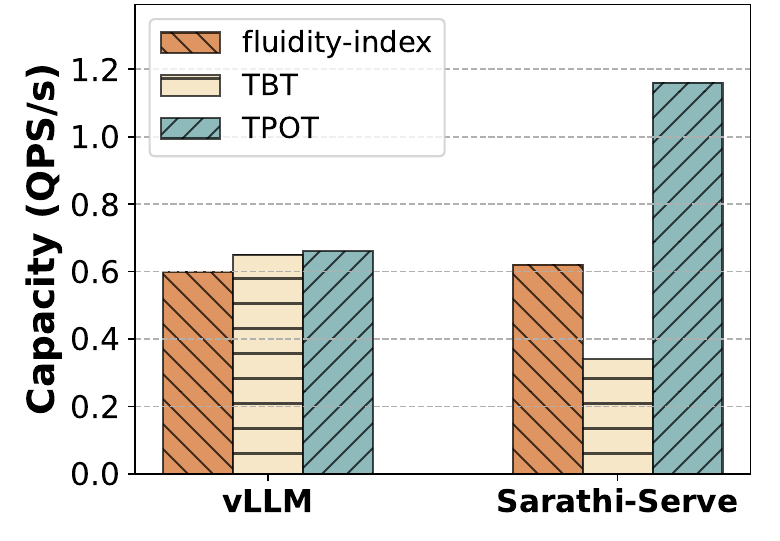} }}%
\label{fig:api:token-rate}
\quad
\subfloat[\small \centering Deadline Miss Rate CDF.]{{\includegraphics[width=0.26\textwidth]{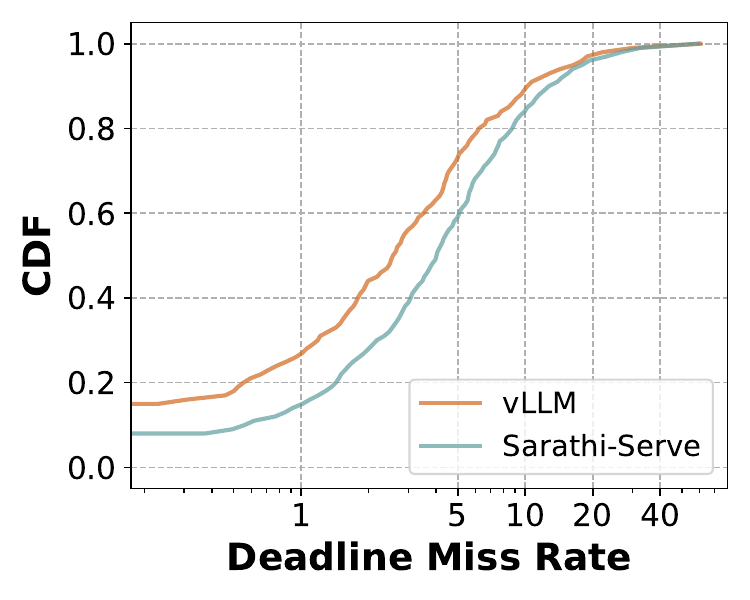} }}%
\label{fig:api:tbt-latency}

\caption{\small Capacity evaluation of open source systems vLLM and Sarathi-Serve performed on H100 for Llama3-8B. (a) shows the overall capacity achieved obtained by using different decode lat. metrics - TBT, TPOT, and \metric. (b) captures the distribution of deadline miss rate at capacity point for the two systems.}
\label{fig:eval:oss}
\end{figure}

\textbf{Results}. Earlier in ~\sref{sec:motivation:challenges}
we compared the vLLM and Sarathi-Serve at a  high load, where we observed that the tail TBT-based throughput for vLLM is 3\myx worse than Sarathi due to huge generation stalls, while the TPOT based throughput shows these systems at par. \metric shows the true difference in throughput between these systems. 
Next, we compare these systems under a strict TBT SLO. We use \metric to define the service SLO as -- 99\% of requests should have less than 10\% deadline miss rate with 25ms target TBT. \sysname then finds the maximum request load (QPS) at which the SLOs can be maintained. We also consider TPOT and P99 TBT based SLOs as baselines with the target latency of 25ms. Here, Sarathi has a 2x lower token throughput compared to vLLM as measured by tail TBT-based metric, in contrast to \autoref{fig:mot:tbt}. At a budget of 25ms, almost all the mixed batches created by Sarathi-Serve with prefill chunks violate the latency SLO threshold. ~\autoref{fig:eval:oss}b, shows the distribution of deadline miss rate when both the systems are operating at the capacity. Also, note that in this setting, the \metric based capacity is the same at 0.6 QPS. We find that vLLM has a higher deadline miss rate at lower percentile. This is expected because, vLLM will have to ingest prefills in their entirety from time to time and whenever it does so, the TBT deadline is breached. In contrast, with chunked prefills Sarathi-Serve achieves fewer deadline misses and higher fluidity.

\section{Discussion}
\label{sec:discussion}

\sysname provides an evaluation framework for LLM inference using \metric metric that tracks the missed deadlines per request. We now discuss the challenges that we leave to future work.

The \metric metric requires setting a deadline for every token -- for the first token of every request, this is the target latency for the prefill phase. In this paper we discuss a potential mechanism for selecting prefill latency target based on observing the prefill processing curve across varying prompt sizes. However, picking a deadline for a given prompt length is challenging for proprietary systems as we cannot accurately characterize their prefill performance; the observed prefill time can include scheduling delays which may offset the expected trends in prefill processing. We leave it to future work to explore alternate ways of prefill latency target selection for propreitery system evaluation.

Next, we observe that we need to provide a small scheduling slack in deadline computations as discussed in \sref{sec:design:unified-deadline}. We pick this value based on our empirical observations; we leave it to future work to systematically set a scheduling slack. Finally, open-source systems have various performance tuning knobs; for e.g., chunk size in Sarathi-Serve, block size in vLLM etc. \sysname currently does not explore or auto-tune such parameters; it expects the users to set the configuration parameters while evaluating across two systems.

\section{Related Work}
\label{section:relatedwork}

Machine learning inference systems have been studied extensively over the last decade. TensorFlow-Serving~\cite{tensorflowserving}, Clipper~\cite{clipper}, BatchMaker~\cite{batchmakereurosys}, and Clockwork~\cite{clockwork} propose various caching, placement, and batching strategies to improve general model serving. More recently works including Orca~\cite{orca}, vLLM~\cite{vllmsosp}, Sarathi \cite{sarathi-serve} primarily addresses the dedicated challenges faced in auto-regressive transformer inference using efficient memory management and scheduling. SplitWise, DistServe and TetriInfer ~\cite{patel2023splitwise,distserve2024,tetriinfer} have presented options to disaggregate the prefill and decode phases to eliminate the interference between them.

\section{Conclusion}
\label{sec:conclusion}

Evaluating LLM inference systems is a challenging problem due to the unique characteristics of autoregressive decode process. We presented a detailed analysis of existing evaluation metrics and their pitfalls. To address their shortcomings, we introduce \sysname -- a holistic LLM evaluation framework that instantiates a novel approach comprised of a novel \metric based approach to evaluating LLM inference systems in a user-facing manner. We then show how \sysname can be leveraged to evaluate both open-source and proprietary model serving systems. \sysname is aimed to serve as a standard evaluation suite for LLM inference systems.

\phantomsection
\label{EndOfPaper}
\bibliographystyle{plain}
\bibliography{main}

\end{document}